\pdfoutput=1

\documentclass{article}

\usepackage[dblblindworkshop, final]{neurips_2026}

\workshoptitle{ATTRIB: Attributing Model Behavior at Scale: Data Attribution and Provenance}

\usepackage[utf8]{inputenc}
\usepackage[T1]{fontenc}
\usepackage{float}
\usepackage[hidelinks]{hyperref}
\usepackage{url}

\usepackage{booktabs}
\usepackage{amsfonts}
\usepackage{nicefrac}
\usepackage{microtype}
\usepackage{xcolor}

\usepackage{xspace}
\usepackage{xcolor}
\usepackage{colortbl}
\definecolor{famworked}{HTML}{EF9A9A}
\definecolor{famprimitive}{HTML}{FFD54F}
\definecolor{famverbal}{HTML}{90CAF9}
\definecolor{famcode}{HTML}{A5D6A7}
\definecolor{famweb}{HTML}{CFCFCF}
\definecolor{verdictok}{HTML}{009E73}
\definecolor{verdictbad}{HTML}{D55E00}
\definecolor{significant}{HTML}{90CAF9}
\colorlet{cellok}{verdictok!12}
\colorlet{cellbad}{verdictbad!12}
\colorlet{cellsig}{significant!45}
\newcommand{\tablestyle}{\footnotesize\renewcommand{\arraystretch}{1.12}\setlength{\tabcolsep}{5pt}}
\newcommand{\tablehead}[1]{\textbf{#1}}
\newcommand{\famtitle}[1]{{\scriptsize\bfseries #1\par}\vspace{1.5pt}}
\newcommand{\famrec}[1]{{\ttfamily\fontsize{7}{8}\selectfont #1\par}\vspace{2pt}}
\newcommand{\famdesc}[1]{{\fontsize{7}{8}\selectfont\itshape #1\par}}

\usepackage{pifont}
\usepackage{enumitem}
\usepackage{etoolbox}
\usepackage{amsmath,amssymb}
\usepackage{graphicx}
\usepackage{tikz}
\usetikzlibrary{arrows.meta,positioning,calc}
\usepackage{twemojis}
\usepackage{placeins}

\newcommand{\reffig}[1]{Figure~\ref{fig:#1}}
\newcommand{\reftab}[1]{Table~\ref{tab:#1}}
\newcommand{\refsec}[1]{Section~\ref{sec:#1}}

\newcommand{\refapp}[1]{Appendix~\ref{app:#1}}

\newcommand{\concept}[1]{\textbf{#1}}
\newcommand{\term}[1]{\textit{#1}}
\newcommand{\code}[1]{\texttt{#1}}

\newif\ifanon
\anonfalse

\newcommand{\emj}[1]{\raisebox{-.12ex}{\twemoji[height=1.05\fontcharht\font`X]{#1}}}
\newcommand{\affil}[1]{\textsuperscript{\normalfont #1}}

\newif\ifdraftmode
\draftmodefalse

\title{MathaMAGICal: Validating Trajectory Data Attribution During Capability Acquisition}

\author{%
  \bfseries
  \href{https://openreview.net/profile?id=~Jason_Lee12}{Jason Lee}\affil{\emj{rocket}\,$*$} \quad
  \href{https://openreview.net/profile?id=~Dikshanta_Pokhrel1}{Dikshanta Pokhrel}\affil{\emj{rocket}\,$*$} \quad
  \href{https://openreview.net/profile?id=~Phonn_Nay_Khant1}{Phonn Nay Khant}\affil{\emj{rocket}\,$*$} \quad
  \href{https://openreview.net/profile?id=~Victor_Lee3}{Victor Lee}\affil{\emj{rocket}\,\emj{bear}\,$*$} \\[2pt]
  \bfseries
  \href{https://openreview.net/profile?id=~Moritz_Weckbecker1}{Moritz Weckbecker}\affil{\emj{rocket}\,\emj{high voltage}} \quad
  \href{https://openreview.net/profile?id=~Sebastian_Lapuschkin1}{Sebastian Lapuschkin}\affil{\emj{high voltage}} \quad
  \href{https://openreview.net/profile?id=~Wojciech_Samek1}{Wojciech Samek}\affil{\emj{high voltage}\,\emj{classical building}} \\[2pt]
  \bfseries
  \href{https://openreview.net/profile?id=~Mark_Riedl1}{Mark Riedl}\affil{\emj{bee}} \quad
  \href{https://openreview.net/profile?id=~Glenn_Matlin1}{Glenn Matlin}\affil{\emj{rocket}\,\emj{bee}\,$\dagger$} \\[4pt]
  \textnormal{\small
    \emj{rocket}\,Algoverse AI Research\hspace{0.8em}%
    \emj{bear}\,University of California, Los Angeles\hspace{0.8em}%
    \emj{high voltage}\,Fraunhofer Heinrich Hertz Institute} \\
  \textnormal{\small
    \emj{classical building}\,Technische Universit\"at Berlin\hspace{0.8em}%
    \emj{bee}\,Georgia Institute of Technology} \\[2pt]
  \textnormal{\footnotesize $^{*}$Equal contribution. \quad $^{\dagger}$Corresponding author: \texttt{glenn@gatech.edu}}
}

\begin{document}

\maketitle

\begin{abstract}

Influence-function-based methods like MAGIC provide exact first-order scores for trajectory data attribution, yet only retraining indicates whether those scores accurately predict data removal. We evaluate MAGIC on $86$M- and $152$M-parameter GPT-2-style models learning multistep arithmetic from scratch on filtered web text and four synthetic math families. Full-run attribution at $86$M suffers from numerical overflow, and even with metagradient clipping, scores correlate near chance with retraining (linear datamodeling score, LDS, $0.17$--$0.22$). In contrast, short checkpoint windows yield stable, accurate predictions at both scales (LDS $0.86$--$0.99$ across four $100$-step windows at $152$M), whereas a single-checkpoint gradient-dot baseline shows near-zero correlation. Rankings of individual training blocks predict retraining only within that single run and do not transfer across seeds or hyperparameters ($\rho \le 0.09$), whereas format-level patterns replicate across runs. Two data families, worked examples and primitive facts, occupy $2.1$--$2.6\times$ their corpus share in both the helpful and harmful $5\%$ score tails, reflecting strong bidirectional local sensitivity distinct from net benefit under multi-epoch family deletion. These results show that trajectory attribution is reliable within short, validated windows and yields family-level insights that generalize across runs.

\end{abstract}

\colorlet{pathfail}{verdictbad}
\colorlet{pathok}{verdictok}
\colorlet{webgray}{famweb}
\colorlet{famred}{famworked}
\colorlet{famyel}{famprimitive}
\colorlet{famblu}{famverbal}
\colorlet{famgrn}{famcode}

\begingroup\setlength{\intextsep}{4pt}
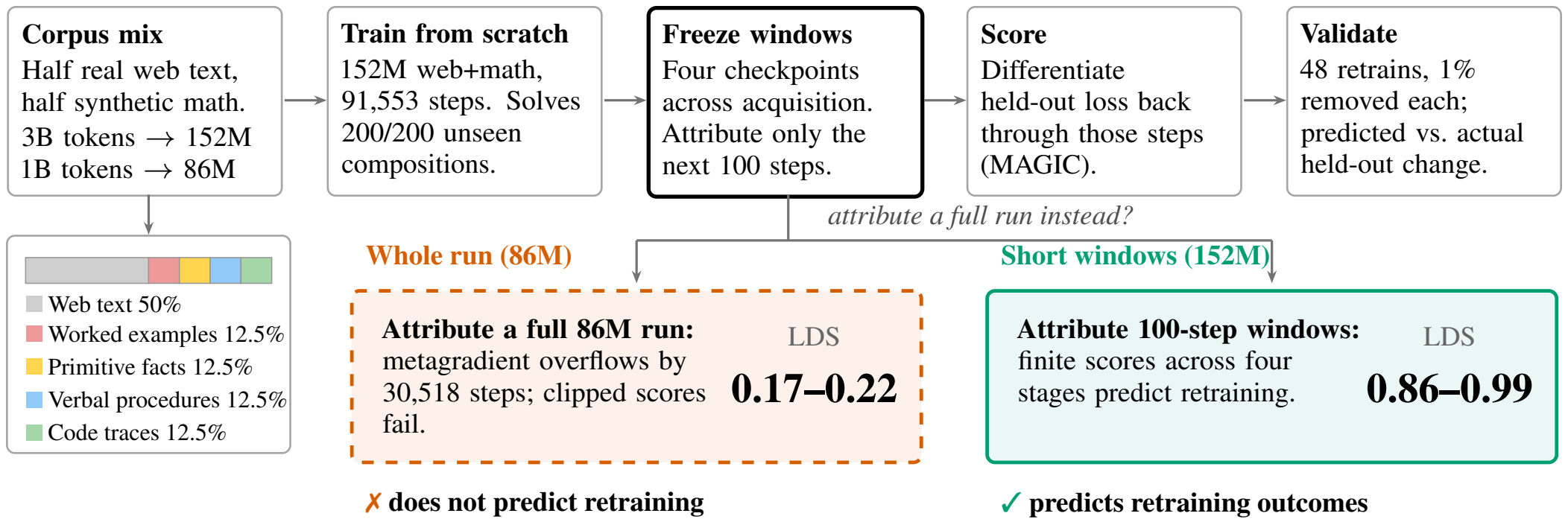
\begin{figure}[H]
  \setlength{\abovecaptionskip}{3pt}\setlength{\belowcaptionskip}{0pt}
  \centering
  \hyphenpenalty=10000\exhyphenpenalty=10000
  \begin{tikzpicture}[x=1in, y=1in, font=\scriptsize,
      card/.style={draw=black!35, line width=0.5pt, rounded corners=2pt, fill=white,
                   text width=0.86in, align=left, inner sep=3.5pt,
                   anchor=north west, minimum height=0.66in},
      anchorcard/.style={card, draw=black, line width=1pt},
      shell/.style={rounded corners=2pt, fill=white, line width=0.9pt,
                    anchor=north, minimum width=1.99in, minimum height=0.60in},
      famshell/.style={draw=black!35, line width=0.5pt, rounded corners=2pt,
                       fill=white, anchor=north, minimum width=0.99in,
                       minimum height=0.76in},
      flow/.style={-{Stealth[length=3.2pt]}, draw=black!55, line width=0.6pt},
      lab/.style={font=\scriptsize\bfseries, anchor=south west}]

    \node[card] (c1) at (0,0) {{\bfseries Corpus mix}\\[1pt]
      Half real web text, half synthetic math.\\[1pt]
      3B tokens $\to$ 152M\\
      1B tokens $\to$ 86M};
    \node[card] (c2) at (1.1175,0) {{\bfseries Train from scratch}\\[1pt]
      152M web+math, 91,553 steps. Solves 200/200 unseen compositions.};
    \node[anchorcard] (c3) at (2.235,0) {{\bfseries Freeze windows}\\[1pt]
      Four checkpoints across acquisition. Attribute only the next 100 steps.};
    \node[card] (c4) at (3.3525,0) {{\bfseries Score}\\[1pt]
      Differentiate held-out loss back through those steps (MAGIC).};
    \node[card] (c5) at (4.47,0) {{\bfseries Validate}\\[1pt]
      48 retrains, 1\% removed each; predicted vs.\ actual held-out change.};

    \foreach \a/\b in {0.97/1.1175, 2.0875/2.235, 3.205/3.3525, 4.3225/4.47}
      \draw[flow] (\a,-0.33) -- (\b,-0.33);

    \draw[flow] (0.495,-0.66) -- (0.495,-0.80);
    \node[famshell] (fam) at (0.495,-0.80) {};
    \begin{scope}[line width=0.3pt]
      \draw[fill=webgray,  draw=black!40] (0.065,-0.88) rectangle +(0.430,-0.09);
      \draw[fill=famred, draw=black!40] (0.495,-0.88) rectangle +(0.1075,-0.09);
      \draw[fill=famyel, draw=black!40] (0.6025,-0.88) rectangle +(0.1075,-0.09);
      \draw[fill=famblu, draw=black!40] (0.710,-0.88) rectangle +(0.1075,-0.09);
      \draw[fill=famgrn, draw=black!40] (0.8175,-0.88) rectangle +(0.1075,-0.09);
    \end{scope}
    \node[anchor=north west, font=\fontsize{6}{6.8}\selectfont, inner sep=0pt] at (0.065,-1.01)
      {\textcolor{webgray}{\rule{4pt}{4pt}}~Web text 50\%};
    \node[anchor=north west, font=\fontsize{6}{6.8}\selectfont, inner sep=0pt] at (0.065,-1.115)
      {\textcolor{famred}{\rule{4pt}{4pt}}~Worked examples 12.5\%};
    \node[anchor=north west, font=\fontsize{6}{6.8}\selectfont, inner sep=0pt] at (0.065,-1.23)
      {\textcolor{famyel}{\rule{4pt}{4pt}}~Primitive facts 12.5\%};
    \node[anchor=north west, font=\fontsize{6}{6.8}\selectfont, inner sep=0pt] at (0.065,-1.345)
      {\textcolor{famblu}{\rule{4pt}{4pt}}~Verbal procedures 12.5\%};
    \node[anchor=north west, font=\fontsize{6}{6.8}\selectfont, inner sep=0pt] at (0.065,-1.46)
      {\textcolor{famgrn}{\rule{4pt}{4pt}}~Code traces 12.5\%};

    \draw[draw=black!55, line width=0.6pt] (2.73,-0.66) -- (2.73,-0.82);
    \draw[draw=black!55, line width=0.6pt] (2.20,-0.82) -- (4.42,-0.82);
    \draw[flow] (2.20,-0.82) -- (2.20,-0.97);
    \draw[flow] (4.42,-0.82) -- (4.42,-0.97);
    \node[anchor=west, font=\scriptsize\itshape, text=black!70]
      at (2.82,-0.74) {attribute a full run instead?};

    \node[lab, text=pathfail] at (1.205,-0.97) {Whole run (86M)};
    \node[shell, draw=pathfail, dashed, fill=pathfail!8] (l2) at (2.20,-0.99) {};
    \node[anchor=north west, text width=1.18in, align=left, font=\scriptsize]
      at ($(l2.north west)+(0.065,-0.055)$)
      {{\bfseries Attribute a full 86M run:} metagradient overflows by
       30,518 steps; clipped scores fail.};
    \node[anchor=center, font=\fontsize{7}{8}\selectfont, text=black!60] at ($(l2.east)+(-0.38,0.14)$) {LDS};
    \node[anchor=center, font=\fontsize{10.5}{11.5}\selectfont\bfseries] at ($(l2.east)+(-0.38,-0.04)$) {0.17--0.22};
    \node[anchor=north west, font=\scriptsize] at (1.205,-1.65)
      {\textcolor{pathfail}{\ding{55}}~{\bfseries does not predict retraining}};

    \node[lab, text=pathok] at (3.425,-0.97) {Short windows (152M)};
    \node[shell, draw=pathok, fill=pathok!8] (r2) at (4.42,-0.99) {};
    \node[anchor=north west, text width=1.18in, align=left, font=\scriptsize]
      at ($(r2.north west)+(0.065,-0.055)$)
      {{\bfseries Attribute 100-step windows:} finite scores across four
       stages predict retraining.};
    \node[anchor=center, font=\fontsize{7}{8}\selectfont, text=black!60] at ($(r2.east)+(-0.38,0.14)$) {LDS};
    \node[anchor=center, font=\fontsize{10.5}{11.5}\selectfont\bfseries] at ($(r2.east)+(-0.38,-0.04)$) {0.86--0.99};
    \node[anchor=north west, font=\scriptsize] at (3.425,-1.65)
      {\textcolor{pathok}{\ding{51}}~{\bfseries predicts retraining outcomes}};
  \end{tikzpicture}
  \caption{Models train from scratch on $50\%$ web text and $50\%$ synthetic
  arithmetic across four data families (bar, left); 152M web+math solves all
  $200/200$ held-out compositions. Differentiating held-out loss through a full
  $86$M run overflows and clipped scores fail retraining validation (LDS
  $0.17$--$0.22$), whereas short checkpoint windows validate at both scales (LDS
  $0.99$ on $25$-step $86$M windows; $0.86$--$0.99$ across four $100$-step $152$M
  windows).}
  \label{fig:overview}
\end{figure}
\endgroup

\section{Introduction}
\label{sec:introduction}

Trajectory-based data attribution methods like MAGIC estimate how training
examples influence model behavior by differentiating held-out loss back through
executed optimizer updates \citep{ilyas2025magic,engstrom2025optimizing}. MAGIC
yields an exact first-order derivative along one training path, yet unrolling
across thousands of pretraining steps overflows numerically, and stabilized
finite scores still require retraining checks to verify whether they predict
data removal. To test when trajectory attribution holds during capability
acquisition without pretraining contamination, we train $86$M and $152$M
GPT-2-style models \citep{radford2019language} from scratch on filtered
FineWeb-Edu text \citep{penedo2024fineweb} and four synthetic arithmetic
families: worked examples, primitive facts, verbal procedures, and code traces
(\reffig{overview}). Models train on single operations and MULTIPLY-then-ADD
chains and are evaluated on held-out ADD-then-MULTIPLY problems (e.g., take
$7$, add $45$, then multiply by $3$ to get $156$). Using \texttt{bergson} \citep{quirke2026bergson}, we test full-run and
short-window MAGIC scores against subset-removal retraining (\refapp{magic}).

\paragraph{Contributions.}
\begin{itemize}[leftmargin=1.2em, itemsep=1pt, topsep=2pt]
  \item \textbf{Horizon limits and short-window validity.} 

    Full-run MAGIC attribution overflows on our $86$M model, and even with
    reverse-pass metagradient clipping, scores correlate near chance with
    retraining (linear datamodeling score, LDS, $0.17$--$0.22$). Short
    continuations from intermediate checkpoints validate at both scales (LDS
    $0.86$--$0.99$ across four $100$-step windows at $152$M), whereas a
    single-checkpoint gradient-dot baseline sits near zero.

  \item \textbf{Family-level sensitivity vs.\ trajectory-bound block rankings.}
    Individual block rankings predict retraining on their own trajectory yet
    fail to transfer across seeds, batch sizes, or optimizers on identical
    blocks (rank correlation $\le 0.09$), whereas family-level patterns
    replicate across runs. Worked examples and primitive facts occupy
    $2.1$--$2.6\times$ their corpus share in both helpful and harmful $5\%$
    score tails, capturing bidirectional local sensitivity distinct from net
    benefit under multi-epoch family deletion.
\end{itemize}

\section{Related Work}
\label{sec:related}

Unlike corroborative text matching \citep{worledge2024unifying,liu2025olmotrace},
contributive attribution estimates counterfactual influence from checkpoint
gradients or subset regressions: influence functions
\citep{koh2017understanding,grosse2023studying}, TracIn
\citep{pruthi2020estimating}, Datamodels \citep{ilyas2022datamodels}, TRAK
\citep{park2023trak}, LESS \citep{xia2024less}, and TrackStar
\citep{chang2025scalable}. MAGIC and metagradient descent differentiate
optimizer updates end-to-end \citep{ilyas2025magic,engstrom2025optimizing} in
\texttt{bergson} \citep{quirke2026bergson}, though horizon drift, cross-run
variance, stagewise sign flips, and group interactions limit retraining
agreement
\citep{nguyen2023bayesian,deng2024dattri,lee2025stagewise,heo2026interaction,deng2026faithful}
and circuit tracing targets internal units \citep{chen2026mechanistic}.

Pretraining studies use attribution or procedural mixtures to shape reasoning
\citep{ruis2024procedural,schoenegger2025influence,kou2025attributes,humane2025influence,rathi2026shaping,jiang2026procedural},
while Transformer arithmetic shows format-sensitive transitions and
compositional limits
\citep{nogueira2021investigating,nanda2023progress,schaeffer2023emergent,dziri2023faith,lee2023teaching,sun2024phases,lumbantobing2026arithmetic}.
Under synthetic provenance control, we test the horizon and granularity where
trajectory attribution predicts retraining (\refapp{extended}).

\section{Method}
\label{sec:method}

To study skill acquisition, we train models from scratch on a mixture of web
text and synthetic arithmetic from four data families (\reffig{data-families}),
built from ADD, SUB, and MULTIPLY operations. During training, we include single
operations and MULTIPLY-then-ADD sequences, but deliberately exclude
ADD-then-MULTIPLY programs. We hold out $200$ ADD-then-MULTIPLY problems as our
test set and score them by exact match. Control models trained without arithmetic
solve none of the $200$ (\refapp{index}).

\colorlet{famred}{famworked}
\colorlet{famyel}{famprimitive}
\colorlet{famblu}{famverbal}
\colorlet{famgrn}{famcode}

\begin{figure}[t]
  \setlength{\abovecaptionskip}{3pt}
  \centering
  \begin{tikzpicture}[x=1in, y=1in,
      fam/.style={rounded corners=2pt, draw=black!30, line width=0.5pt,
                  text width=1.15in, align=left, inner sep=4pt,
                  anchor=north west, minimum height=0.95in}]
    \node[fam, fill=famred!55] at (0,0)
      {\famtitle{Worked examples}
       \famrec{START: 17 | STEP\_1: MULTIPLY 2 | STEP\_2: ADD 6 | FINAL: ?
       >{}> WORK: STEP\_1: 34 | STEP\_2: 40 | FINAL: 40}
       \famdesc{the evaluation format}};
    \node[fam, fill=famblu!55] at (1.41,0)
      {\famtitle{Verbal procedures}
       \famrec{Start with 17. Multiply by 2 to get 34. Add 6 to get 40.}
       \famdesc{the same execution in natural language}};
    \node[fam, fill=famgrn!55] at (2.82,0)
      {\famtitle{Code traces}
       \famrec{x = 17\\ x = x * 2 \# x = 34\\ x = x + 6 \# x = 40}
       \famdesc{the same execution as an assignment-style trace}};
    \node[fam, fill=famyel!55] at (4.23,0)
      {\famtitle{Primitive facts}
       \famrec{88 * 2 = 176\\ 412 + 6 = 418}
       \famdesc{independent operands: the operations, never their composition}};
  \end{tikzpicture}
  \caption{Schematic family formats for the training-allowed order MULTIPLY
  then ADD. The first three show the same execution; primitive facts use
  independent operands. Actual program identifiers are exclusive to one
  family per seed (Appendix~\ref{app:corpus}).}
  \label{fig:data-families}
\end{figure}

\paragraph{Attribution and retraining validation.} We target the negative
log-likelihood of held-out output tokens and use \texttt{bergson}
\citep{quirke2026bergson} to differentiate this loss through executed training
steps with MAGIC \citep{ilyas2025magic}, assigning one metagradient score per
$512$-token block. Each score estimates the first-order effect of removing a
block: positive scores mark harmful blocks (removal lowers held-out loss) and
negative scores mark helpful ones, matching retrained loss changes at $r = +0.99$
on validated windows (\refapp{attribution}). To validate predictions, we remove
sampled block subsets, retrain from the same checkpoint, and compute the linear
datamodeling score (LDS) \citep{ilyas2022datamodels,deng2024dattri}, the rank
correlation between predicted and observed loss changes per held-out query ($1$
is exact rank agreement; $0$ is chance).

\paragraph{Two sets of metrics.} Because per-block rankings do not transfer
across independent runs (\refsec{families}), we evaluate data families with two
complementary metrics. \emph{Tail lift} measures local sensitivity as a family's
enrichment in the most helpful and most harmful $5\%$ of scored blocks relative
to its corpus share; equal enrichment in both tails indicates high variance
whose opposing effects largely cancel in the mean. \emph{Net benefit} measures
macro-scale reliance via multi-epoch deletion of an entire family
(\refsec{corroboration}).

\begin{figure}[t]
  \centering
  \includegraphics[width=\linewidth]{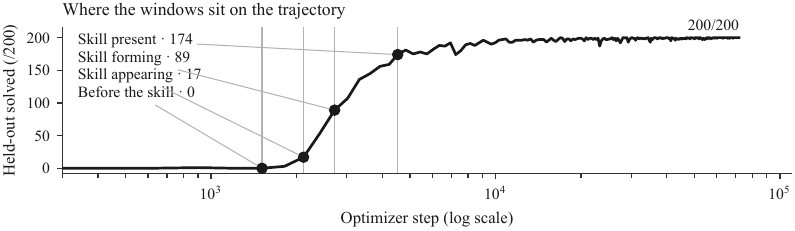}
  \caption{Held-out accuracy of 152M web+math. Each window attributes $100$
  steps from its marked checkpoint; the model reaches $200/200$ by a third of
  training and holds it.}
  \label{fig:acquisition-152m}
\end{figure}

\paragraph{Attribution windows.} Because attributing the full training run
fails validation (\refsec{fullrun}), we attribute short windows instead. Each
window starts from a saved checkpoint and trains for $100$ more steps on new
blocks. We attribute only those $100$ steps, since longer windows stop matching
retraining (\refsec{windows}). We place the four main $152$M windows before the
skill ($0/200$), as it appears ($17/200$), as it forms ($89/200$), and once it
is present ($174/200$) (\reffig{acquisition-152m}; \reftab{window-index}).

\paragraph{Corpus controls.} We match families on token budget and on reuse,
the average number of times each distinct program appears, so no family has
more data or repetition than another. Reuse is $5.0\times$ in the $86$M corpus
(\refapp{corpus}) and $4\times$ in the main $152$M run (\refapp{single-phase}).
Contamination audits of the web text find zero matches to our record formats or
held-out items, and web-only controls solve $0/200$ (\refapp{corpus}).

\subsection{Experimental design}
\label{sec:design}
\label{sec:experiments}

\paragraph{Models.} Models are named by parameter count and training diet
(\refapp{index}):
\emph{web+math} (single mixed run), \emph{web$\to$math} or \emph{code$\to$math}
(arithmetic-free pretrain followed by a math phase), and \emph{web-only} or
\emph{code-only} (controls). The main model is \textbf{152M web+math} ($3.0$B
tokens); two-phase $152$M runs saw fewer arithmetic tokens than the $86$M series
and are flagged where reported (\refapp{extended}).

\paragraph{Stabilizers, seeds, and baselines.} We test stronger damping, forward
gradient clipping, and Muon \citep{jordan2024muon,liu2025muon} (altering forward
training) alongside reverse-pass metagradient clipping (leaving forward weights
unchanged): at $86$M ($189/200$ reference), stronger damping collapses learning
($0/200$), whereas forward clipping ($193/200$) and Muon ($194/200$) preserve
capability (\reftab{damping}). Most windows use one seed; full-run instability uses two and
deletion three. Because reweighting alters the subsequent trajectory, across
which per-block rankings do not transfer (\refsec{families}), we evaluate family
deletion rather than block upweighting. TrackStar and SOURCE
\citep{chang2025scalable,quirke2026bergson} exceeded compute limits, so our
baseline is single-checkpoint gradient-dot on identical retraining subsets.
Code, checkpoints, and scores are released (\refapp{benchmarks}).

\section{Results}
\label{sec:results}

\subsection{Full-run attribution overflows, and clipped scores fail validation}
\label{sec:fullrun}
Full-run attribution overflows at $86$M: the running metagradient grows by
$0.4$--$0.5\%$ per step, leaving $44.7\%$ of block scores non-finite after
$30{,}518$ steps. Clipping the running metagradient in an otherwise
bit-identical rerun eliminates non-finite scores, yet aggregate LDS stays near
chance at $0.217$ and $0.168$ across two seeds ($12$ subsets each;
$p = 0.50, 0.60$), showing that finiteness alone does not ensure retraining
agreement. Stronger optimizer damping likewise fails to preserve both capability
and full-run stability (\reffig{horizon}; \reftab{damping}, Appendix),
motivating short-window attribution.

\subsection{Short-window attribution validates against retraining}
\label{sec:windows}
Across $24$ short attribution windows at $86$M ($23$ with full $1\%$-removal
retraining grids), we vary horizon length, batch size, damping, optimizer,
acquisition stage, seed, and query set. At $512$ rows per step, windows reach
LDS $0.90$--$0.99$; extending the horizon to $400$ steps drops LDS to $-0.08$,
and shrinking batch size to $64$ rows drops it to $0.08$ (step count and batch
size co-vary across fixed-budget windows; \reftab{envelope}, Appendix).

When we use attribution windows before, during, and after emergence, we obtain
LDS values ranging from $0.61$ to $0.93$. This shows continuous loss
sensitivity across short continuations while exact-match accuracy changes by
only a handful of problems before the main increase in capability (\refapp{interface}).

\subsection{Short-window validation across acquisition at 152M}
\label{sec:mainmodel}
The main 152M web+math model reaches $200/200$ held-out and $100/100$
in-distribution exact match one-third through its $3.0$B-token run and stays
saturated (\reffig{acquisition-152m}). On unseen blocks, four $100$-step windows
probe \emph{before the skill}, \emph{skill appearing}, \emph{skill forming}, and
\emph{skill present}. Across $48$ retraining subsets and eight queries per
window, median per-query LDS reaches $\ge 0.86$ in all four windows and
$\ge 0.99$ in three (\reftab{windows-152m}, Appendix), lowest at
\emph{skill appearing}. All four share one parent run, and batch/horizon
differences preclude isolating a clean $86$M-to-$152$M scaling comparison.

\subsection{First-order baseline comparison}
\label{sec:baseline}
Using the same retraining subsets, we apply single-checkpoint gradient-dot
attribution. This achieves near-zero LDS compared to $0.57$--$0.99$ for
unrolled MAGIC (\reffig{baseline}, Appendix). This confirms the value of
short-horizon unrolling.

\subsection{Family-level sensitivity vs.\ trajectory-bound block rankings}
\label{sec:families}
\begin{figure}[t]
  \centering
  \includegraphics[width=\linewidth]{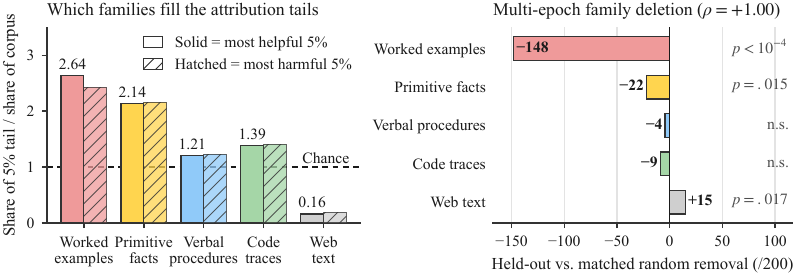}
  \caption{Family-level sensitivity vs.\ multi-epoch deletion on 152M web+math.
  \emph{Left}: both $5\%$ score tails at initial emergence as lift over corpus
  share; worked examples and primitive facts populate both tails (bidirectional
  local sensitivity), while web text stays depleted. \emph{Right}: multi-epoch
  family deletion against step-matched random removal ($18$ retrains, $3$ seeds)
  measures net benefit outside the local regime and aligns in four-family rank
  ($\rho = +1.00$). Worked examples share the evaluation format, and web removal
  shifts the arithmetic fraction.}
  \label{fig:lift-152m}
\end{figure}

Primitive facts and worked examples are enriched in both extreme $5\%$ tails of
validated attribution scores at $2.1$--$2.6$ times their corpus share, while code
traces are mildly enriched, verbal procedures are chance-like, and web text is
less than a quarter of its chance share (\reffig{lift-152m};
\reftab{family-lift-152m}, Appendix). The symmetric enrichment in both tails
shows that attribution detects high bidirectional local sensitivity rather than
uniform helpfulness across a family. Because worked examples share the held-out
evaluation format, the detected sensitivity reflects both arithmetic composition
and alignment with the evaluation format.

Individual block rankings predict retraining on their own trajectory but do not
transfer across runs. On the same set of blocks, across different seeds, batch
sizes, or optimizers, block-level rankings of sensitivity fail to generalize
(Spearman correlations between $-0.01$ and $+0.09$; top-$5\%$ overlap near
chance; \reftab{block-agreement}, Appendix). In contrast, the family-level tail
enrichment measures are reproducible across these same changes, establishing
that data families are the reproducible unit across trajectories.

\subsection{Stage dynamics of family tail sensitivity}
\label{sec:dynamics}
\begin{figure}[t]
  \centering
  \includegraphics[width=\linewidth]{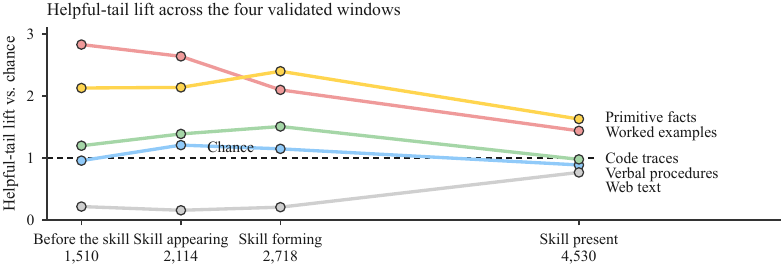}
  \caption{Helpful-tail lift per family across the four validated $152$M windows,
  plotted at their starting checkpoints (four discrete windows; line segments aid
  visual tracking only). Harmful-tail lifts follow nearly identical trajectories
  (\reftab{family-lift-152m}, Appendix), reflecting symmetric bidirectional local
  sensitivity.}
  \label{fig:family-lift-stages}
\end{figure}

\reffig{family-lift-stages} illustrates how the enrichment of the helpful
tail changes over the four $152$M attribution windows. Worked examples are most
enriched before accuracy emerges and primitive facts are most enriched during
emergence. Web text is less than a quarter of its chance share until the model
saturates, at which point the family-level shares approach the corpus
proportions. The harmful tail shows the same pattern of movement
(\reftab{family-lift-152m}, Appendix). The per-block absolute effect sizes
diminish as the held-out loss drops, whereas the loss-normalized sensitivity
metric increases and then saturates (\reftab{effect-sizes-152m}, Appendix).

\subsection{Multi-epoch family deletion}
\label{sec:corroboration}
To measure net family benefit independently of local tail sensitivity, we run
$18$ continuations of 152M web+math from \emph{skill appearing}: one deletion
arm per family plus a step-matched random-deletion control, each removing
$3{,}200$ of $25{,}600$ blocks over eight epochs ($3$ seeds per arm).

Deleting worked examples halts learning ($0/200$ held-out accuracy vs.\
$148/200$ under random removal). Deleting primitive facts also reduces accuracy
($126/200$), while deleting code traces ($139/200$) or verbal procedures
($143.7/200$) stays near random (\reffig{lift-152m}; \reftab{deletion-152m},
Appendix). Four-family deletion loss aligns in rank ($\rho = +1.00$) with window
tail lift despite operating outside the local linear regime. Removing web text
raises the arithmetic share from $50.0\%$ (random control) to $57.1\%$,
confounding its $+15.0$-problem gain.

\subsection{Two-phase pretraining and checkpoint conditioning}
\label{sec:scale}
Pretraining without arithmetic alters subsequent acquisition speed: across a
matched $800$-step math phase at $86$M, code$\rightarrow$math reaches $130/200$
held-out exact match versus $26/200$ for web$\rightarrow$math. Attribution
windows within both math phases validate at LDS $0.995$ and $0.996$; window
tails emphasize worked examples while math-phase deletion relies on worked
examples and primitive facts, and deleting code traces costs no more than random
removal (\refapp{extended}). Pretraining thus alters readiness to learn, whereas
math-phase attribution captures local sensitivity within that phase.

\section{Discussion}
\label{sec:discussion}
\label{sec:limitations}

Window attribution measures how strongly individual blocks perturb held-out loss
along a single trajectory, whereas multi-epoch deletion tests macro-scale
reliance when an entire family is withheld. Worked examples and primitive facts
populate both the most helpful and most harmful $5\%$ tails while keeping
near-zero mean scores, reflecting wide block-to-block variance rather than
uniformly positive updates. Because worked examples match the fixed evaluation
format, their sensitivity captures both arithmetic composition and prompt
formatting.

\paragraph{Limitations.}
Most window configurations use one seed, restricting cross-run claims to the
family level: per-block rankings are trajectory-specific and do not transfer. In
the $86$M grid, batch size and step count co-vary, and web deletion raises the
arithmetic share. We make no cross-scale capability claims: two-phase $152$M
runs saw fewer math tokens than $86$M runs, and 152M web+math uses a single
parent seed under a different window configuration. Arithmetic accuracy holds
only within the trained numeric range, and TrackStar and SOURCE exceeded our
compute budget (\refapp{extended}).

\section{Conclusion}
\label{sec:conclusion}

Short-window MAGIC attribution predicts retraining upon removing data during
capability acquisition, but full-run unrolling overflows and clipped full-run
attribution fails validation. Within-run rankings are trajectory-specific and
fail to transfer, but the family-level tail patterns replicate across runs.
These patterns represent local bidirectional sensitivity, which is different
from the multi-epoch net impact. Thus, trajectory attribution is most useful
when used within a validated window and interpreted as a family-level effect.

\label{sec:acknowledgements}

\begin{ack}
This research was supported in part through research cyberinfrastructure resources and services provided by the Partnership for an Advanced Computing Environment (PACE) at the Georgia Institute of Technology, Atlanta, Georgia, USA. RRID:SCR\_027619. We also thank \href{https://modal.com/}{Modal Labs} for additional compute support and infrastructure that accelerated our retraining validation experiments, and Algoverse AI Research for research mentorship, program support, and compute resources. Finally, we thank Lucia Quirke, Louis Jaburi, Gon\c{c}alo Paulo, Stella Biderman, and the EleutherAI team for open-source trajectory-attribution tooling in \texttt{bergson} and helpful technical discussions. The authors declare no competing financial interests.
\end{ack}

\bibliographystyle{plainnat}
\bibliography{references}

\clearpage
\appendix
\section{Author Contributions}
\label{app:contributions}

Jason Lee, Dikshanta Pokhrel, Phonn Nay Khant, and Victor Lee contributed equally as joint first authors; Glenn Matlin served as senior and corresponding author. The matrix below records each author's roles under the NISO CRediT (Contributor Roles Taxonomy) standard.

{%
\definecolor{creditFill}{HTML}{1F4E79}%
\definecolor{creditStroke}{HTML}{0F2A43}%
\definecolor{creditEmpty}{HTML}{B8B8B8}%
\newcommand{\crY}{\tikz[baseline=-0.55ex]\filldraw[fill=creditFill, draw=creditStroke, line width=0.35pt] (0,0) circle (2.3pt);}%
\newcommand{\crN}{\tikz[baseline=-0.55ex]\fill[creditEmpty] (0,0) circle (0.85pt);}%
\newcommand{\crH}[1]{\multicolumn{1}{c}{\rotatebox{90}{\scriptsize #1}}}%
\begin{center}
\footnotesize
\setlength{\tabcolsep}{0pt}
\renewcommand{\arraystretch}{1.22}
\begin{tabular}{@{}l@{\hspace{1.3em}}*{14}{>{\centering\arraybackslash}p{2.05em}}@{}}
  & \crH{Conceptualization} & \crH{Data curation} & \crH{Formal analysis}
  & \crH{Funding acquisition} & \crH{Investigation} & \crH{Methodology}
  & \crH{Project administration} & \crH{Resources} & \crH{Software}
  & \crH{Supervision} & \crH{Validation} & \crH{Visualization}
  & \crH{Writing -- original draft} & \crH{Writing -- review \& editing} \\
\addlinespace[3pt]
\midrule
Jason Lee              & \crY & \crY & \crN & \crN & \crY & \crY & \crN & \crY & \crY & \crN & \crY & \crY & \crY & \crY \\
Dikshanta Pokhrel      & \crY & \crN & \crY & \crN & \crY & \crY & \crN & \crN & \crY & \crN & \crY & \crN & \crY & \crY \\
Phonn Nay Khant        & \crN & \crY & \crN & \crN & \crY & \crY & \crN & \crN & \crY & \crN & \crY & \crY & \crY & \crY \\
Victor Lee             & \crN & \crN & \crN & \crN & \crY & \crY & \crN & \crN & \crY & \crN & \crY & \crY & \crY & \crY \\
\addlinespace[4pt]
Moritz Weckbecker      & \crN & \crN & \crN & \crN & \crN & \crY & \crN & \crN & \crN & \crY & \crN & \crN & \crN & \crN \\
\addlinespace[4pt]
Sebastian Lapuschkin   & \crN & \crN & \crN & \crN & \crN & \crN & \crN & \crN & \crN & \crY & \crN & \crN & \crN & \crN \\
Wojciech Samek         & \crN & \crN & \crN & \crN & \crN & \crN & \crN & \crN & \crN & \crY & \crN & \crN & \crN & \crN \\
Mark Riedl             & \crN & \crN & \crN & \crN & \crN & \crN & \crN & \crN & \crN & \crY & \crN & \crN & \crN & \crN \\
\addlinespace[4pt]
Glenn Matlin           & \crY & \crN & \crN & \crY & \crY & \crY & \crY & \crY & \crY & \crY & \crN & \crY & \crY & \crY \\
\bottomrule
\end{tabular}
\end{center}
}

\section{Additional Details}
\label{app:additional}

\subsection{Corpus specification}
\label{app:corpus}

A \term{program} is a starting value with an ordered list of
operations, where each operation is an operator paired with an operand.
The example in \reffig{data-families} starts at $17$, multiplies by $2$, and
then adds $6$.
\reftab{program-space} gives the allowed ranges and the five compositions, or
two-step operator orders, used in training.
If any intermediate or final value falls outside the allowed range, we discard
the program and draw a new one.
There are $1{,}000$ start values, $99$ addition or subtraction operands, and
$11$ multipliers.
Each of the three compositions that include a multiplication therefore allows
$1{,}000 \times 99 \times 11 = 1{,}089{,}000$ programs, and each of the two that
use only addition and subtraction allows
$1{,}000 \times 99 \times 99 = 9{,}801{,}000$, for $2.29\times10^{7}$ in all.
Discarding programs whose value drops below zero leaves $2.22\times10^{7}$.
Our main 152M run raises the largest start value to $2{,}999$
(\refapp{single-phase}).

\begin{table}[ht]
  \centering
  \caption{Program space of the 86M specification. A program whose intermediate or
  final value leaves $[0,\,99{,}999]$ is rejected and resampled. Under these
  ranges no program reaches the ceiling, so only the floor rejects. ADD followed
  by MULTIPLY is withheld for evaluation.}
  \label{tab:program-space}
  \tablestyle
  \setlength{\tabcolsep}{6pt}
  \begin{tabular}{@{}ll@{\hskip 2em}ll@{}}
    \toprule
    \tablehead{Parameter} & \tablehead{Range} & \tablehead{Composition} & \tablehead{Operator order} \\
    \midrule
    Start value        & $[0,\,999]$        & \code{mul\_add} & MUL, ADD \\
    ADD/SUB operand    & $[1,\,99]$         & \code{add\_sub} & ADD, SUB \\
    Multiplier         & $[2,\,12]$         & \code{sub\_add} & SUB, ADD \\
    Value floor        & $0$                & \code{sub\_mul} & SUB, MUL \\
    Value ceiling      & $99{,}999$         & \code{mul\_sub} & MUL, SUB \\
    \midrule
    Distinct programs  & $2.22\times10^{7}$ & \code{add\_mul} & \emph{Held out} \\
    \bottomrule
  \end{tabular}
\end{table}

\concept{Families.}
We render programs in four families at equal token budgets
(\reffig{data-families}).
Worked examples, verbal procedures, and code traces write out a whole program
step by step, so each operation acts on the result of the one before and every
intermediate value appears.
Primitive facts take each operation from a program and apply it to a newly
drawn starting value instead of the previous result: for the example program,
$88 \times 2 = 176$ and $412 + 6 = 418$, where a worked example has
$17 \times 2 = 34$ and $34 + 6 = 40$.
The model sees the same single operations but never sees one result feed into
the next.
Within a seed, each program appears in only one family, and each family has a
balanced mix of the five compositions.
Worked examples are the only family written in the evaluation format, which
states the problem and then a \code{WORK} marker.
At evaluation, the model must generate every number after that marker itself.

\concept{Token budgets and reuse.}
The families render a program at different lengths (\reftab{record-lengths}), so
equal problem counts would give them unequal shares of the corpus.
We equalize budgets in tokens.
Reusing programs raises held-out loss and makes scores larger and less stable
(\reftab{reuse-cost}), and one distinct-program count for all families would
leave reuse confounded with family: an earlier build that did this ranged from
$3.9\times$ to $14.1\times$.
We therefore set each family's distinct-program count to its token budget divided
by the product of the target reuse and its mean record length.
The 1B-token corpus has a common reuse of $5.0\times$, with $232{,}000$ distinct
programs for worked examples and $845{,}000$ for primitive facts.
The main 152M corpus targets a reuse of four (\refapp{single-phase}).

\concept{Score scale.}
Family tail statistics rank raw block scores.
Packed synthetic blocks carry slight trailing padding relative to full
$512$-token web blocks (e.g., $503$ vs.\ $512$ content tokens for primitive
facts); normalizing scores per content token on the $86$M runs leaves family
lifts virtually unchanged.

\concept{Identifier audit and web screening.}
Program identifiers are truncated content hashes of the program signature: the
start value, operator order, and operands.
We audit every build by checking that all records sharing an identifier also
share a signature. The 1B-token corpus has no collisions across $1{,}954{,}000$
identifiers, and the 3B-token corpus has none across $7{,}076{,}856$.
Held-out problems come from the same generator and ranges as training, so we
audit all $103{,}966$ web documents ($500$\,MB) in the 1B-token corpus against
our record format, every held-out prompt, and every held-out answer fact: zero
documents match any of the three. Arithmetic-free web-only controls at both
$86$M and $152$M ($1.0$B web tokens) likewise solve $0/200$ held-out items
(\reftab{model-index}).
The $200$ held-out problems are frozen across every model we report.

\concept{Builds.}
Builds are deterministic given a seed.
Each build writes a manifest with token counts, block counts, realized reuse,
distinct-program counts, the identifier audit, and a hash over the packed corpus.
The generator, packer, tokenizer, and mixture manifest are released with each
corpus.

\begin{table}[t]
  \centering
  \caption{Mean tokens per record for each family. These lengths set the
  distinct-program count each family needs to reach a common reuse at an equal
  token budget.}
  \label{tab:record-lengths}
  \tablestyle
  \begin{tabular}{@{}lr@{}}
    \toprule
    \tablehead{Family} & \tablehead{Mean tokens per record} \\
    \midrule
    Worked examples      & $107.6$ \\
    Primitive facts      & $29.6$ \\
    Verbal procedures    & $62.9$ \\
    Code traces          & $52.1$ \\
    \bottomrule
  \end{tabular}
\end{table}

\begin{table}[t]
  \centering
  \caption{Reuse degrades capability at a matched token budget. Attribution scores
  also grew in magnitude and agreed less between runs as reuse rose.}
  \label{tab:reuse-cost}
  \tablestyle
  \begin{tabular}{@{}rrr@{}}
    \toprule
    \tablehead{Distinct programs} & \tablehead{Reuse} & \tablehead{Held-out loss} \\
    \midrule
    $1{,}500$ & $\sim7\times$   & $2.47$ \\
    $200$     & $\sim48\times$  & $2.66$ \\
    $50$      & $\sim173\times$ & $2.76$ \\
    \bottomrule
  \end{tabular}
\end{table}

\FloatBarrier

\subsection{Interface and format sensitivity}
\label{app:interface}

Held-out exact match evaluates one held-out composition in one fixed prompt
format. Two broader evaluations test how far the trained models generalize
across prompt interfaces and operand magnitudes across all fifteen models of
\refapp{benchmarks}.

\begin{figure}[t]
  \centering
  \includegraphics[width=\linewidth]{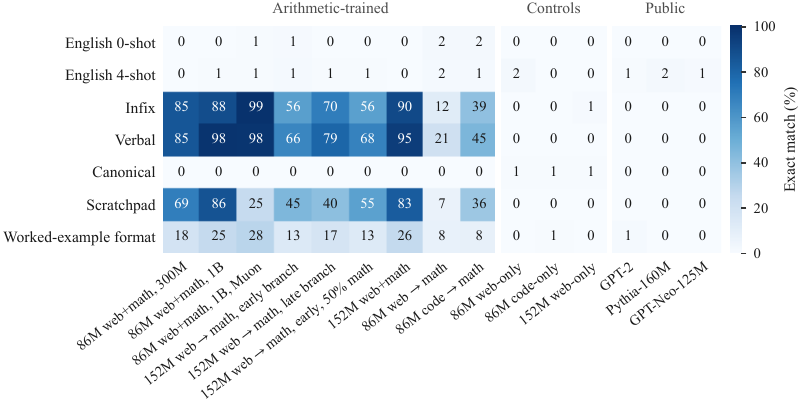}
  \caption{Exact match by model and prompt format on the $783$ benchmark problems
  inside the 86M training specification ($788$ for infix). Columns are nine
  arithmetic-trained models, three arithmetic-free controls, and three public
  models (GPT-2, Pythia-160M, GPT-Neo-125M).}
  \label{fig:format-heatmap}
\end{figure}

\paragraph{Interface benchmark.}
We pose $2{,}000$ problems from EleutherAI's \texttt{asdiv} and
\texttt{arithmetic} suites
\citep{gao2021evalharness,miao2020diverse,brown2020language} in seven prompt
formats; \reffig{format-heatmap} reports the $783$ problems inside the 86M
training specification. Arithmetic-trained models reach up to $99\%$ in infix
or verbal form and at most $2\%$ in English on the identical underlying
problems, showing that they compute reliably within trained syntax without
comprehending natural-language problem statements. The arithmetic-free and
public controls stay below $2\%$ in every format, and every arithmetic-trained
model scores $0\%$ in \texttt{canonical} form (answer only, without intermediate
steps) because generation reverts to the worked-solution template seen during
training.

\paragraph{Discrete exact match vs.\ continuous completion loss.}
We select acquisition stages by discrete exact match rather than continuous
completion loss because completion loss already approaches a digit-guessing
floor while exact match remains $0/200$ \citep{schaeffer2023emergent}. MAGIC
instead differentiates continuous completion loss, which explains why
pre-emergence windows at $0/200$ exact match still carry informative gradients
and validate at LDS $0.61$--$0.991$ (\refsec{windows}). Because accuracy swings
from $99\%$ to near $0\%$ across prompt interfaces above, the main attribution
experiments hold one evaluation format fixed across stages and seeds so that
measured sensitivities reflect arithmetic learning rather than prompt mismatch
(\refsec{discussion}).

\begin{figure}[t]
  \centering
  \includegraphics[width=\linewidth]{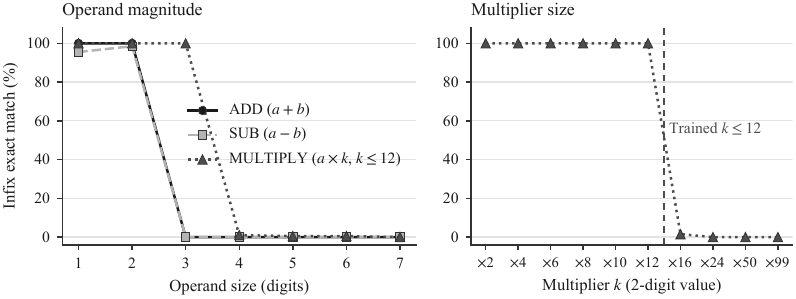}
  \caption{Exact match on generated single-operation infix problems by operand size
  (left) and multiplier (right), for the strongest 86M model (86M web+math, 1B,
  Muon). Accuracy holds up to the trained limits of two-digit addition and
  subtraction operands, three-digit multiplicands, and multipliers of twelve, and
  drops to near zero beyond them.}
  \label{fig:numrange-cliffs}
\end{figure}

\paragraph{Number range.}
A second evaluation sweeps single-operation infix problems across operand size
(\reffig{numrange-cliffs}). The arithmetic-trained models stop at the edge of their
training range. None exceeds $0.5\%$ on addition or subtraction with three-digit
operands, and none exceeds $4\%$ at a multiplier of $16$. On four-digit multiplicands, all but one stay at or below $4\%$. The exception
is 152M web+math at $47.5\%$ ($95/200$) before dropping to $0\%$ at five digits:
its specification raises the maximum start value from $999$ to $2{,}999$,
placing $49$ of the $200$ ($24.5\%$) four-digit items inside its training range
and indicating partial generalization across the remainder of the four-digit
bucket. The two-phase 152M models keep the 86M specification and stop at the
same limits as the 86M models.

\FloatBarrier

\subsection{Attribution and validation detail}
\label{app:attribution}

\begin{figure}[t]
  \centering
  \includegraphics[width=\linewidth]{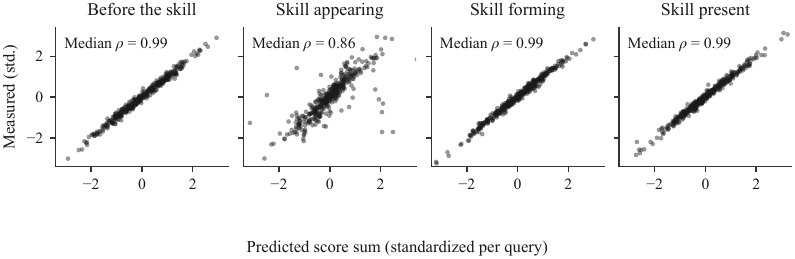}
  \caption{Predicted against measured effect for every leave-$1\%$-out retrain
  of the four 152M web+math windows ($48$ subsets $\times$ $8$ queries each,
  standardized per query). Each panel prints the median per-query LDS.}
  \label{fig:validation-scatter}
\end{figure}

\paragraph{Sign check.}
For each removed subset, the validation harness records baseline held-out loss
minus the loss after removal, and correlates that difference with the subset's
summed scores. On the four main windows, the correlation is positive for all $32$
query-window pairs, with a median of $0.99$ in three windows and $0.86$ in the
window where accuracy first appears (\reffig{validation-scatter}). A positive
score therefore marks a block whose removal lowers held-out loss.

The family analyses read scores from disk, so we also regenerated the $48$ subsets
of each main window from their seed and summed the stored scores. The sums match
the harness's records on $384$ of $384$ rows in every window.

\paragraph{Effect sizes.}
We measure the size of a window's retraining effects by the standard deviation of
the held-out loss change across its $48$ removals (\reftab{effect-sizes-152m}).
This standard deviation falls by a factor of $44$ from the first 152M window to the
last, which could suggest that training data stops mattering once the skill is
present. Baseline held-out loss falls by a factor of about $600$ over the same
windows, so relative to the remaining loss the effect rises from $0.08\%$ to about
$1.1\%$ by the third window and then plateaus. The smallest standard deviation we report,
$6\times10^{-6}$, is above the fp32 evaluation precision of about $10^{-6}$.

\paragraph{Artifact availability.} The window checkpoints, per-block score
tensors, $48\times8$ retraining grids, deletion indices, and the base run's
evaluation curve are released as a results dump alongside the checkpoints and
corpora (\refapp{benchmarks}).

\begin{table}[t]
  \centering
  \caption{Retraining effect sizes across the four 152M windows. SD is the standard
  deviation of the held-out loss change across the $48$ removals of $1\%$ of
  blocks. Baseline loss and SD are medians over the eight queries.}
  \label{tab:effect-sizes-152m}
  \tablestyle
  \begin{tabular}{@{}lrrr@{}}
    \toprule
    \tablehead{Window} & \tablehead{Baseline loss} & \shortstack{\tablehead{Median SD}\\\tablehead{($\boldsymbol{\Delta}$loss)}} & \tablehead{SD / baseline} \\
    \midrule
    Before the skill & $0.342$   & $2.66\times10^{-4}$ & $0.0008$ \\
    Skill appearing  & $0.101$   & $2.27\times10^{-4}$ & $0.0022$ \\
    Skill forming    & $0.0067$  & $7.67\times10^{-5}$ & $0.0114$ \\
    Skill present    & $0.00055$ & $6.09\times10^{-6}$ & $0.0111$ \\
    \bottomrule
  \end{tabular}
\end{table}

\begin{figure}[t]
  \centering
  \includegraphics[width=\linewidth]{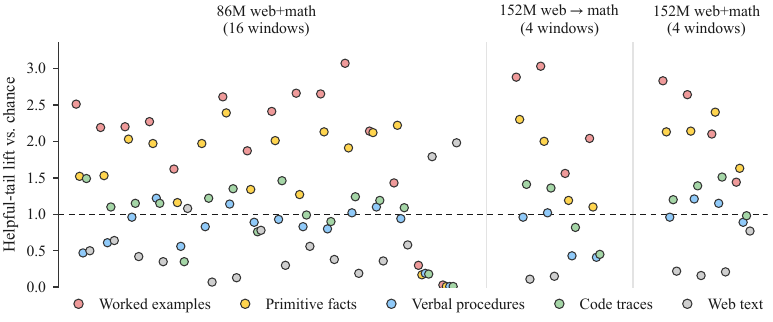}
  \caption{Helpful-tail lift per family for every window with a full family
  table in the released results ($24$ windows across three model groups). The
  ranking of Figure~\ref{fig:lift-152m} replicates across configurations and
  scales. The 86M windows where web text rises and worked examples fall are on
  fully trained models, where every math block had been seen and web blocks were
  fresh; that data-freshness artifact is discussed with the limitations.}
  \label{fig:window-grid}
\end{figure}

\begin{figure}[t]
  \centering
  \includegraphics[width=0.62\linewidth]{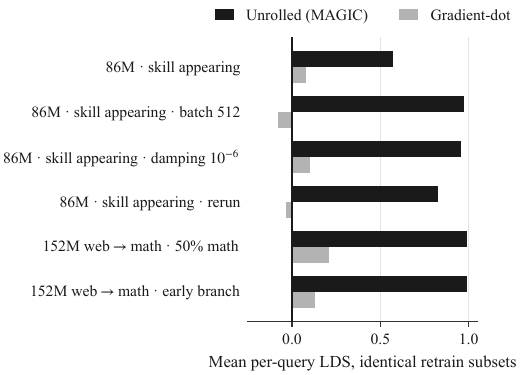}
  \caption{Aggregate LDS for the unrolled scores and the first-order gradient-dot baseline,
  computed on the identical retrain subsets, for four windows at 86M and two at 152M.}
  \label{fig:baseline}
\end{figure}

\FloatBarrier

\subsection{Single-phase run protocol}
\label{app:single-phase}
\concept{Base run.}
152M web+math is a $152$M-parameter decoder with context $512$ and a
$276$-token byte-and-digit tokenizer, trained for one epoch over $3.0$B tokens
at a $50/50$ mixture of web text and synthetic arithmetic with a target reuse of
four.
Optimization uses AdamW \citep{loshchilov2019decoupled} at learning rate
$3\times10^{-4}$ with linear decay, weight decay $0.01$, $20$ warmup steps,
batch $64$, and $91{,}553$ optimizer steps.
The program specification raises the maximum start value to $2{,}999$, which
takes the rarest composition to $3.21$M programs and the space to approximately $6.8\times10^{7}$.

\concept{Window protocol.}
All four windows use an identical protocol: $25{,}600$ blocks for two epochs at
batch $512$ with gradient accumulation of eight, giving $100$ optimizer steps; a
math fraction of $0.5$; damping of $10^{-6}$; AdamW at a constant learning rate
of $2\times10^{-4}$ with five warmup steps; full single precision; evaluation
mode; eight held-out queries; and $48$ leave-one-percent-out retrains.
Continuation blocks are drawn from the portion of the corpus the base run had not
reached in its own shuffle order.

\concept{Checkpoint resume.}
The base run reached its platform wall-clock limit at step $88{,}788$ ($97.0\%$),
resumed deterministically from the stored parameters, optimizer state, batch
index, and random-number state, and completed the remaining $2{,}765$ steps.
Checkpoint restoration itself is constant-time because training restarts directly
at the saved batch index without replaying prior data; however, each container
launch rebuilds the tokenized dataset on startup ($\approx 2.3$~hours even with a
persistent volume cache), so finishing the final $3\%$ required $2.35$~hours of
startup preparation for $0.7$~hours of training.

\subsection{Model and window index}
\label{app:index}
Models are named by parameter count and training diet. \emph{web+math} is a model
trained once, from scratch, on the mixed corpus; \emph{web$\to$math} and
\emph{code$\to$math} are pretrained without arithmetic and then given a math phase;
\emph{web-only} and \emph{code-only} never see arithmetic and are the floor
controls. A qualifier appears only where one is needed to separate two runs, so a
math phase is $100\%$ math unless stated otherwise, and \emph{early} and \emph{late
branch} are the halfway and final checkpoints of the $152$M web-only pretrain.
All models in the $152$M family have $152$M parameters; we never write $150$M.

Windows are named \emph{model $\cdot$ stage}, where the four stages are fixed by
held-out accuracy at the window's initialization: \emph{before the skill} at
$0/200$, \emph{skill appearing} at the first non-zero score, \emph{skill forming}
on the rise, and \emph{skill present} near ceiling.

The validity-envelope sweeps of Section~\ref{sec:windows} are all variants of the
86M web+math, 300M $\cdot$ skill appearing window, and carry the run IDs
\texttt{A1-len200}, \texttt{A1-len400} and \texttt{A1-len800} for horizon,
\texttt{A3-bs64}, \texttt{A3-bs128} and \texttt{A3-bs512} for batch,
\texttt{A4-muon} for the optimizer, \texttt{A5-eps0} and \texttt{A5-eps1e6} for
damping, and \texttt{A7-indist} for the in-distribution target. Deletion arms are
named \texttt{C1v2-L-drop-\textit{family}-s\textit{n}} over three seeds.

Full per-window configuration and results for all $44$ windows are in the released
results dump.

\begin{table}[ht]
  \centering
  \caption{Model index. Every model named in the paper, with the subfolder that
  backs it in \texttt{leejason2025/mathamagical-checkpoints} on Hugging Face. The run name used in
  our code and logs is the middle segment of each subfolder. Held-out is
  free-running exact match on the $200$ ADD${\to}$MULTIPLY problems.}
  \label{tab:model-index}
  \tablestyle
  \setlength{\tabcolsep}{4pt}
  \begin{tabular}{@{}llrr@{}}
    \toprule
    \tablehead{Paper name} & \tablehead{HF subfolder} & \tablehead{Tokens} & \tablehead{Held-out} \\
    \midrule
    \textbf{152M web+math} & \texttt{150m-mix150m-3b/step\_91552} & $3.0$B & $\mathbf{200/200}$ \\
    \addlinespace[2pt]
    86M web+math, 1B & \texttt{86m-cap2-s0/step\_30517} & $1.0$B & $189/200$ \\
    86M web+math, 1B, Muon & \texttt{86m-cap2-muon-s0/step\_30517} & $1.0$B & $194/200$ \\
    86M web+math, 300M & \texttt{86m-cap2-300m-clip-s0/step\_9155} & $300$M & $177/200$ \\
    \addlinespace[2pt]
    86M web$\to$math & \texttt{86m-CUR-nl-mathphase/step\_800} & $300$M${+}800$ & $26/200$ \\
    86M code$\to$math & \texttt{86m-CUR-code-mathphase/step\_800} & $300$M${+}800$ & $130/200$ \\
    152M web$\to$math, late branch & \texttt{150m-P150s-b30517-m1.0/step\_800} & $1.0$B${+}800$ & $141/200$ \\
    152M web$\to$math, early branch & \texttt{150m-P150s-b15312-m1.0/step\_800} & $0.5$B${+}800$ & $129/200$ \\
    \quad Early branch, $50\%$ math & \texttt{150m-P150s-b15312-m0.5/step\_1600} & $0.5$B${+}1{,}600$ & $136/200$ \\
    \addlinespace[2pt]
    86M web-only & \texttt{86m-nl86m-300m/step\_9155} & $300$M & $0/200$ \\
    86M code-only & \texttt{86m-code86m-300m/step\_9155} & $300$M & $0/200$ \\
    152M web-only & \texttt{150m-web150m-1b/step\_30517} & $1.0$B & $0/200$ \\
    \bottomrule
  \end{tabular}
\end{table}

\begin{table}[t]
  \centering
  \caption{Window index. Each attribution window continues training from an
  initialization checkpoint and attributes only the steps it runs. Stage names are
  fixed by held-out accuracy at initialization. Configuration is optimizer steps
  $\times$ rows per step.}
  \label{tab:window-index}
  \tablestyle
  \setlength{\tabcolsep}{4pt}
  \begin{tabular}{@{}lllr@{}}
    \toprule
    \tablehead{Window} & \tablehead{Run ID} & \tablehead{Initialized from} & \tablehead{Configuration} \\
    \midrule
    \multicolumn{4}{@{}l}{\emph{152M web+math}} \\
    \quad Before the skill & \texttt{W3B-before} & \texttt{step\_1510} & $100 \times 512$ \\
    \quad Skill appearing & \texttt{W3B-onset} & \texttt{step\_2114} & $100 \times 512$ \\
    \quad Skill forming & \texttt{W3B-midrise} & \texttt{step\_2718} & $100 \times 512$ \\
    \quad Skill present & \texttt{W3B-after} & \texttt{step\_4530} & $100 \times 512$ \\
    \addlinespace[3pt]
    \multicolumn{4}{@{}l}{\emph{86M web+math, 300M}} \\
    \quad Before the skill & \texttt{winB2-2375} & \texttt{step\_2375} & $100 \times 256$ \\
    \quad Skill appearing & \texttt{winB2-2755} & \texttt{step\_2755} & $100 \times 256$ \\
    \quad Skill appearing, seed 1 & \texttt{winB2-2755-s1} & \texttt{step\_2755} & $100 \times 256$ \\
    \quad Skill forming & \texttt{winB2-3135} & \texttt{step\_3135} & $100 \times 256$ \\
    \quad Skill present & \texttt{C2-post5700} & \texttt{step\_5700} & $50 \times 256$ \\
    \addlinespace[3pt]
    \multicolumn{4}{@{}l}{\emph{152M web$\to$math}} \\
    \quad Late branch & \texttt{W150s-b30517-m1.0-L100} & \texttt{step\_400} & $100 \times 512$ \\
    \quad Early branch & \texttt{W150s-b15312-m1.0-L100} & \texttt{step\_400} & $100 \times 512$ \\
    \quad Early branch, $50\%$ math & \texttt{W150s-b15312-m0.5-L100} & \texttt{step\_700} & $100 \times 512$ \\
    \addlinespace[3pt]
    \multicolumn{4}{@{}l}{\emph{Curriculum arms}} \\
    \quad 86M web$\to$math & \texttt{CUR-nl-win} & \texttt{step\_700} & $25 \times 512$ \\
    \quad 86M code$\to$math & \texttt{CUR-code-win} & \texttt{step\_400} & $25 \times 512$ \\
    \bottomrule
  \end{tabular}
\end{table}

\FloatBarrier

\subsection{Evaluation protocol}
\label{app:benchmarks}
\concept{Models.}
The interface and range evaluations cover fifteen models: twelve of our own
(nine arithmetic-trained models across $86$M and $152$M, including the two
curriculum arms, and the three arithmetic-free controls 86M web-only, 86M
code-only, and 152M web-only), together with three public models of comparable
size: GPT-2 \citep{radford2019language}, Pythia-160M \citep{biderman2023pythia},
and GPT-Neo-125M \citep{black2021gptneo}.
One further checkpoint is omitted deliberately: it is a bit-identical forward twin
of another model in the set.

\concept{Interface benchmark.}
The same $2{,}000$ arithmetic problems are presented in seven formats: English
zero-shot and four-shot, numeric infix, verbal, canonical, scratchpad, and the
specification-then-work format the models train on.
Decoding is greedy, and scoring is exact match.
Results are reported both over all items and restricted to items inside each
model's training specification, because the two differ substantially.

\concept{Number-size range evaluation.}
Single-operation problems are generated and swept across operand magnitude and
multiplier size to locate where each model's arithmetic fails.
The sweep covers addition and subtraction from one to seven-digit operands, and
multiplication across multiplicand digits and multipliers from two to
ninety-nine.

\concept{Released artifacts.}
All public artifacts are hosted at three repositories:
\begin{itemize}[nosep, leftmargin=*, before=\raggedright]
  \item \textbf{Model checkpoints:} \url{https://huggingface.co/leejason2025/mathamagical-checkpoints}
  \item \textbf{$3.0$B training corpus:} \url{https://huggingface.co/datasets/leejason2025/mathamagical-3b-data}
  \item \textbf{Code and validation summaries:} \url{https://github.com/HCAI-Lab-GT/mathamagical}
\end{itemize}
Reproducing the main $152$M results uses the released base and four window
checkpoints, the exact training corpus and frozen held-out set, the corpus
manifest and hash, and the production generator, packer, and tokenizer that
ship with the corpus.

Independent validation uses the released results dump: per-query
correlations, the $48\times8$ retraining grid for each main window, score
tensors, subset identities, and run configurations. Family-deletion results
ship as seed-level outputs with the matched random-deletion controls, and the
base evaluation curve identifies the selected acquisition stages. Every chart
in this paper is generated from those files.

\subsection{Additional results and limitations}
\label{app:extended}

\paragraph{Baselines.}
We planned to compare against TrackStar and SOURCE
\citep{chang2025scalable,quirke2026bergson} on the same runs, targets, and
queries. Neither produced scores within our compute budget: SOURCE exceeded a
$24$-hour cap while preconditioning its first segment, and TrackStar was
mis-scoped and not rerun. Stronger checkpoint-based estimators therefore remain
untested.

In their place we use single-checkpoint gradient-dot, which needs no unrolling.
At each window's starting checkpoint, it scores every block by the dot product
between the block's gradient and the held-out loss gradient. We validate these
scores on the same retraining subsets as the MAGIC scores. Gradient-dot reaches
LDS $0.08$, $-0.08$, $0.10$, and $-0.04$ on four $86$M windows and $0.21$ and
$0.13$ on two $152$M windows, against $0.57$--$0.99$ for MAGIC on those subsets
(\reffig{baseline}). Per-block Spearman correlation between the two methods is
$0.00$--$0.16$. Gradient-dot also puts arithmetic in the helpful tail and web
text in the harmful tail, which looks plausible, but its scores do not agree
with retraining.

\begin{figure}[t]
  \centering
  \includegraphics[width=\linewidth]{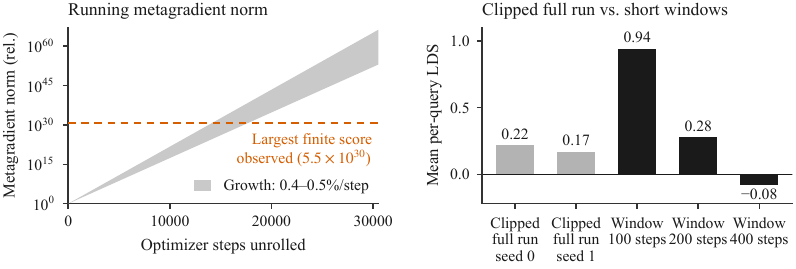}
  \caption{\emph{Left}: growth of the running metagradient norm at the measured rate
  of $0.4$--$0.5\%$ per step, against the largest finite score observed.
  \emph{Right}: mean per-query LDS against retraining for the clipped full-run
  scores (two seeds) and for short windows of $100$, $200$, and $400$ steps.
  Clipping removes the overflow without restoring agreement; shorter windows
  restore it.}
  \label{fig:horizon}
\end{figure}

\paragraph{Full-run attribution.}
The \texttt{bergson} implementation of MAGIC carries the running metagradient
backward through every optimizer step with no clipping, normalization, or
damping, and its norm grows by about $0.4$--$0.5\%$ per step (\reffig{horizon},
left). At $30{,}518$ steps, $44.7\%$ of block scores are non-finite and the
largest finite magnitude is $5.5 \times 10^{30}$. We patched the reverse pass to
clip the running metagradient and reran a bit-identical twin of an affected run.
Non-finite scores fall from $25.3\%$ on the unpatched run to $0\%$, but the
clipped scores do not predict retraining: aggregate LDS is $0.217$ and $0.168$ on
two seeds ($12$ subsets; statistical $p$-values $0.50$ and $0.60$), and
per-query values vary by $\pm 0.3$ within a single run, consistent with
horizon-induced Taylor drift in trajectory attribution \citep{deng2026faithful}.

The library also exposes a damping term, $\epsilon_{\mathrm{root}}$. No setting
in our sweep both learned arithmetic and gave stable full-run scores
(\reftab{damping}). Whether a different optimizer or reverse-pass treatment can
do so remains open.

\begin{table}[ht]
  \centering
  \caption{Held-out exact match out of $200$ across forward-training recipes at
  $86$M ($1.0$B tokens; reverse-pass metagradient clipping leaves forward
  weights unchanged). A model at $0/200$ has no arithmetic to attribute, so the
  damped configurations are excluded. Green marks capability preserved and
  orange capability destroyed.}
  \label{tab:damping}
  \tablestyle
  \begin{tabular}{@{}lr@{}}
    \toprule
    \tablehead{Recipe} & \tablehead{Held-out (/200)} \\
    \midrule
    Conventional reference & $189$ \\
    Muon optimizer & \cellcolor{cellok}$\mathbf{194}$ \\
    Forward gradient clipping ($\|\nabla\| \le 1$) & \cellcolor{cellok}$193$ \\
    Damping $\epsilon_{\mathrm{root}} = 10^{-4}$ or $10^{-2}$ & \cellcolor{cellbad}$0$ \\
    \bottomrule
  \end{tabular}
\end{table}

\paragraph{Windows.}
Across the $86$M sweep, LDS spans $-0.08$ to $0.996$; \reftab{envelope} gives
the main settings. Batch size and horizon are confounded, because steps equal
blocks times epochs divided by batch, and every $86$M window at $512$ rows per
step is only $25$ steps long. A $128$-row window at $100$ steps scores $0.854$,
above the $0.570$ of a $256$-row window at the same length, so we read steps and
batch jointly and name neither as the main factor. The $86$M run at $512$ rows
and $100$ steps that would separate them was not run.

We scored the end-of-window checkpoints to see what changes inside a window.
Held-out accuracy moves by a few problems out of $200$ at both scales (at most
$1 \to 8$), and not at all in the $25$-step windows at $512$ rows. The windows
attribute a decrease in loss; the accuracy jump comes later.

Three windows on the two-phase $152$M models validate at $0.990$, $0.993$, and
$0.994$, with every query above $0.985$, and five more reach $0.94$--$0.99$ on
incomplete retraining banks. The configuration therefore works at a second model
size. We make no cross-scale claim from this, since these models saw $26.2$M
arithmetic tokens against $500$M for the 1B-token $86$M models.

\begin{table}[t]
  \centering
  \caption{Measured validity envelope for short-window attribution at $86$M parameters, one
  axis at a time from the settled configuration ($512$ rows per optimizer step,
  $\epsilon_{\mathrm{root}} = 10^{-6}$, AdamW, constant lr $2\times10^{-4}$, fp32).
  Mean per-query LDS against retraining. Steps and batch are confounded across the grid and
  should be read jointly. Green cells mark LDS $\geq 0.8$, orange cells mark LDS
  $\leq 0.3$; intermediate values are left untinted.}
  \label{tab:envelope}
  \tablestyle
  \begin{tabular}{@{}llr@{}}
    \toprule
    \tablehead{Axis} & \tablehead{Setting} & \tablehead{LDS} \\
    \midrule
    Horizon ($256$ rows/step) & $100$ steps & $0.57$--$0.94$ \\
    Horizon ($256$ rows/step) & $200$ steps & \cellcolor{cellbad}$0.28$ \\
    Horizon ($256$ rows/step) & $400$ steps & \cellcolor{cellbad}$-0.08$ \\
    \addlinespace[2pt]
    Batch & $64$ rows/step & \cellcolor{cellbad}$0.08$ \\
    Batch & $128$ rows/step, $100$ steps & \cellcolor{cellok}$0.854$ \\
    Batch & $512$ rows/step & \cellcolor{cellok}$\mathbf{0.97}$--$\mathbf{0.99}$ \\
    \addlinespace[2pt]
    Boundary position & Before the skill & $0.61$ \\
    Boundary position & Skill forming & \cellcolor{cellok}$0.90$ \\
    Boundary position & Skill present & \cellcolor{cellok}$0.93$ \\
    \addlinespace[2pt]
    Damping & $\epsilon_{\mathrm{root}} = 0$ / $10^{-6}$ & Non-finite / $0.96$ \\
    Optimizer & Muon & \cellcolor{cellok}$0.84$ \\
    Target & In-distribution & \cellcolor{cellok}$0.90$ \\
    \bottomrule
  \end{tabular}
\end{table}

\reftab{windows-152m} gives validation for the four 152M web+math windows. The
model reaches $200/200$ by step $30{,}200$ and holds that score through step
$91{,}552$. The $86$M series behaves differently: LDS rises monotonically across
its windows, and in its window before accuracy appears, two of eight queries fail
validation. The two series also differ in batch size and continuation length, so
we report this as a difference between two configurations and make no
cross-scale claim.

\begin{table}[t]
  \centering
  \caption{Attribution windows on 152M web+math. Each window continues from its
  checkpoint on $25{,}600$ unseen blocks for $100$ optimizer steps and attributes
  only those steps. LDS is the per-query rank correlation between the predicted
  and measured effects of removing $1\%$ of blocks, over a $48 \times 8$
  retraining grid. All $32$ query-window cells reach $p < 10^{-10}$.}
  \label{tab:windows-152m}
  \tablestyle
  \begin{tabular}{@{}lrrrl@{}}
    \toprule
    \tablehead{Window} & \tablehead{Init step} & \shortstack{\tablehead{Held-out}\\\tablehead{at init}} & \tablehead{Median LDS} & \tablehead{LDS range} \\
    \midrule
    Before the skill & $1{,}510$ & $0/200$   & $0.991$ & $0.980$--$0.995$ \\
    Skill appearing  & $2{,}114$ & $17/200$  & $0.863$ & $0.777$--$0.960$ \\
    Skill forming    & $2{,}718$ & $89/200$  & $0.992$ & $0.983$--$0.996$ \\
    Skill present    & $4{,}530$ & $174/200$ & $0.991$ & $0.985$--$0.994$ \\
    \bottomrule
  \end{tabular}
\end{table}

\paragraph{Family tails.}
We compare families by tail lift: a family's share of the most helpful or most
harmful $5\%$ of blocks in a window, divided by its share of the window, so that
$1$ is chance (\reftab{family-lift-152m}). Tail lift measures sensitivity, not
net benefit (\refsec{method}): mean per-family scores are small and change sign
between windows. Worked examples and primitive facts have the largest lifts, and
those lifts are nearly equal in the two tails: $2.83$ and $2.63$ for worked
examples and $2.13$ and $2.15$ for primitive facts in the first window. Blocks
from these families move held-out loss strongly in both directions, which does
not make either family helpful as a whole.

Web text is half the corpus by tokens and holds $0.16$--$0.24$ of its chance
share of either tail during acquisition. The $86$M reference window shows the
same depletion ($0.30$--$0.41$; \reftab{lift}) on a corpus where web text was a
minority of tokens. A low tail share does not mean web text contributes nothing
in total. By the last window the lifts move toward $1$ and web text rises to
$0.77$--$0.83$, so strong family concentration belongs to the acquisition
period.

Four windows that share identical blocks and differ only in batch size,
optimizer, or damping agree on block rankings at Spearman $-0.010$ to $+0.091$,
and one window's eight queries agree with each other at $+0.075$ to $+0.418$
(\reftab{block-agreement}). Rankings from different $86$M seeds agree at $0.00$
to $0.08$, and the top-$5\%$ blocks of two runs overlap at about the rate
expected by chance. MAGIC scores describe one realized training run
(\refapp{magic}), so changing a hyperparameter changes what is being estimated;
\citet{nguyen2023bayesian} describe the same cross-run variance in per-example
scores more generally. Family tail shares hold across these changes, and block
rankings do not.

\begin{table}[t]
  \centering
  \caption{Family share of the most helpful and most harmful $5\%$ of blocks, as a
  lift over each family's share of the window; a lift of one is chance. Entries
  are helpful / harmful, averaged over eight queries.}
  \label{tab:family-lift-152m}
  \tablestyle
  \setlength{\tabcolsep}{5pt}
  \begin{tabular}{@{}lccccc@{}}
    \toprule
    \tablehead{Window} & \tablehead{Worked} & \tablehead{Primitive} & \tablehead{Verbal} & \tablehead{Code} & \tablehead{Web text} \\
    \midrule
    Before the skill & $2.83$\,/\,$2.63$ & $2.13$\,/\,$2.15$ & $0.96$\,/\,$1.07$ & $1.20$\,/\,$1.21$ & $0.22$\,/\,$0.23$ \\
    Skill appearing  & $2.64$\,/\,$2.43$ & $2.14$\,/\,$2.16$ & $1.21$\,/\,$1.23$ & $1.39$\,/\,$1.41$ & $0.16$\,/\,$0.19$ \\
    Skill forming    & $2.10$\,/\,$1.92$ & $2.40$\,/\,$2.48$ & $1.15$\,/\,$1.16$ & $1.51$\,/\,$1.48$ & $0.21$\,/\,$0.24$ \\
    Skill present    & $1.44$\,/\,$1.21$ & $1.63$\,/\,$1.61$ & $0.89$\,/\,$0.82$ & $0.98$\,/\,$1.02$ & $0.77$\,/\,$0.83$ \\
    \bottomrule
  \end{tabular}
\end{table}

\begin{table}[t]
  \centering
  \caption{Block-level rank agreement between four windows that share identical
  blocks and differ only in batch size, optimizer, or damping.}
  \label{tab:block-agreement}
  \tablestyle
  \begin{tabular}{@{}lr@{\hskip 2em}lr@{}}
    \toprule
    \tablehead{Pair} & \tablehead{Agreement} & \tablehead{Pair} & \tablehead{Agreement} \\
    \midrule
    Base vs.\ large batch    & $+0.091$ & Base vs.\ Muon           & $+0.005$ \\
    Base vs.\ damped         & $+0.020$ & Large batch vs.\ Muon    & $+0.012$ \\
    Large batch vs.\ damped  & $-0.010$ & Muon vs.\ damped         & $+0.040$ \\
    \bottomrule
  \end{tabular}
\end{table}

\paragraph{Family deletion.}
Tail lift covers $100$ steps of one run. To measure a family's net benefit, we
delete it and retrain (\reftab{deletion-152m}). Eighteen continuations branch
from the checkpoint where accuracy first appears: one arm per family plus a
random-deletion control, with three seeds each. Every arm removes $3{,}200$ of
$25{,}600$ blocks and trains the remaining $22{,}400$ for eight epochs from the
same state. Deleting worked examples eliminates acquisition ($0/200$ against
$148/200$ for random deletion), and deleting primitive facts lowers accuracy to
$126/200$. Deleting code traces ($139/200$) or verbal procedures ($143.7/200$)
stays closer to random. Three of five contrasts are significant at $n = 3$.
Deletion cost ranks the families in the same order as helpful-tail lift
($\rho = +1.00$), although eight epochs of full-family removal lie far outside
the $1\%$, $100$-step additive first-order regime that the scores approximate
\citep{ilyas2025magic,heo2026interaction}. Worked examples also share the
evaluation format (\refsec{families}); primitive facts share neither the format
nor chained operations and still rank second on both measures.

A fixed block count does not remove the same share of each family. Removing
$3{,}200$ blocks deletes all of an arithmetic family ($12.5\%$ of the corpus) but
only a quarter of web text ($50\%$). The four arithmetic arms train at $42.9\%$
arithmetic and are comparable with each other. The web arm trains at $57.1\%$, so
its $15.0$-problem gain over random deletion cannot be attributed to web removal
alone.

\begin{table}[t]
  \centering
  \caption{Multi-epoch family deletion at $152$M. Eighteen continuations branch
  from the checkpoint where accuracy first appears; each removes $3{,}200$ of
  $25{,}600$ blocks and trains the remaining $22{,}400$ blocks for eight epochs
  from the same initialization, matching step count and starting state. For
  reference, the rightmost column lists helpful-tail lift in the corresponding
  $100$-step window (harmful-tail lifts are similarly elevated;
  \reftab{family-lift-152m}). Four-family deletion loss aligns in rank with
  window tail lift ($\rho = +1.00$), though eight-epoch full-family removal
  measures macro-scale net benefit outside the local linear regime. Web removal
  raises the arithmetic share to $57.1\%$, confounding its $+15.0$-problem gain.
  Bold, blue-tinted $p$-values mark contrasts significant at $n=3$, from
  two-sided pooled two-sample $t$-tests against the random control ($\mathrm{df}=4$).}
  \label{tab:deletion-152m}
  \tablestyle
  \begin{tabular}{@{}lrrrrr@{}}
    \toprule
    \tablehead{Removed} & \shortstack{\tablehead{Held-out (/200)}\\\tablehead{Three seeds}} & \tablehead{Mean} & \tablehead{Vs.\ random} & \tablehead{\textit{p}} & \shortstack{\tablehead{Helpful}\\\tablehead{tail lift}} \\
    \midrule
    Worked examples  & $0$, $0$, $0$       & $0.0$   & $-148.0$ & \cellcolor{cellsig}$\mathbf{<0.0001}$ & $2.64$ \\
    Primitive facts  & $119$, $126$, $133$ & $126.0$ & $-22.0$  & \cellcolor{cellsig}$\mathbf{0.015}$   & $2.14$ \\
    Verbal procedures& $142$, $143$, $146$ & $143.7$ & $-4.3$   & $0.318$   & $1.21$ \\
    Code traces      & $137$, $139$, $141$ & $139.0$ & $-9.0$   & $0.076$   & $1.39$ \\
    Web text         & $161$, $163$, $165$ & $163.0$ & $+15.0$  & \cellcolor{cellsig}$\mathbf{0.017}$   & $0.16$ \\
    \midrule
    Random control   & $143$, $146$, $155$ & $148.0$ & ---      & ---       & --- \\
    \bottomrule
  \end{tabular}
\end{table}

The $86$M reference window gives the same direction with less power
(\reftab{lift}). Retraining the $86$M math phase without worked examples costs
$45.3$ problems against matched random removal ($p = 0.01$), removing primitive
facts costs $9.3$ ($p = 0.23$), and removing verbal or web blocks costs nothing.
Three seeds support the ordering and no stronger claim.

\begin{table}[t]
  \centering
  \caption{Window tail sensitivity and family deletion at $86$M. \emph{Lift} is
  each family's share of the most helpful (most negative $5\%$) and most harmful
  (most positive $5\%$) validated block scores in a reference $86$M window,
  averaged over the six of eight queries with $p < 0.05$ and
  divided by corpus share (\refsec{method}); high-sensitivity families populate
  both tails. \emph{Deletion} reports held-out exact match ($/200$, three seeds)
  after retraining the math phase without that family against step-matched random
  removal (base $64$, random $45$). Only worked examples reach significance at
  $n=3$ ($p = 0.01$; primitive facts $p = 0.23$).}
  \label{tab:lift}
  \tablestyle
  \begin{tabular}{@{}lrrrr@{}}
    \toprule
    & \multicolumn{2}{c}{Lift vs.\ chance} & \multicolumn{2}{c}{Deletion} \\
    \cmidrule(lr){2-3}\cmidrule(lr){4-5}
    \tablehead{Family} & \tablehead{Helpful} & \tablehead{Harmful} & \tablehead{Held-out} & \tablehead{Vs.\ random} \\
    \midrule
    Worked examples & $\mathbf{2.41}$ & $1.85$ & $0$ & $\mathbf{-45.3}$ \\
    Primitive facts & $2.01$ & $\mathbf{2.07}$ & $36$ & $-9.3$ \\
    Verbal procedures & $0.93$ & $1.21$ & $47$ & $+1.7$ \\
    Code traces & $1.46$ & $1.22$ & $43$ & $-2.7$ \\
    Web text & $0.30$ & $0.41$ & $52$ & $+7.0$ \\
    \bottomrule
  \end{tabular}
\end{table}

\paragraph{Curriculum arms.}
86M web$\to$math and 86M code$\to$math differ only in pretraining corpus; their
$800$-step math phase is identical. After it, code$\to$math scores $130/200$
against $26/200$ for web$\to$math (\reftab{model-index}). Because a window has
fixed length, attribution cost does not grow with the length of pretraining.
The two arms' math-phase windows validate at LDS $0.995$ and $0.996$
(\reftab{window-index}), with worked examples enriched in both $5\%$ tails
($2.3$--$2.6\times$ helpful; $2.0$--$2.6\times$ harmful). In $600$-step
family-deletion continuations from each arm's pre-transition checkpoint (two
seeds per arm; random-removal means $129.5/200$ on web$\to$math and $160.5/200$
on code$\to$math), removing worked examples reduces accuracy to $0/200$ in both
arms and removing primitive facts lowers it to $113.0/200$ ($-16.5$) and
$153.5/200$ ($-7.0$), whereas removing code traces ($135.0/200$ and $160.0/200$)
or verbal procedures ($140.5/200$ and $170.5/200$) costs no more than random
removal.

\paragraph{Limitations.}
The results concern one synthetic arithmetic specification, one withheld
operation order, one evaluation format, and one background web corpus.
Contamination audits rule out direct matches to our record formats or held-out
items without eliminating general arithmetic in web text; the arithmetic-free
controls score at most $2\%$ in any prompt format (\reffig{format-heatmap}).

On the fully trained $86$M models, web text rises and worked examples fall in the
helpful tail (\reffig{window-grid}). Every math block in those windows had
already been seen in training while the web blocks were new, so we read this as a
data-freshness artifact and not as a change in the family ordering.

\FloatBarrier

\subsection{MAGIC and influence functions}
\label{app:magic}

This appendix states the parts of \citet{ilyas2025magic} that our claims depend
on and applies them to our target, source unit, and sign convention
(\refsec{method}). \citet{ilyas2025magic} give the full development and the
exactness argument.

\paragraph{The influence function.}
Represent a dataset as a weight vector $\mathbf{w} \in \mathbb{R}^{n}$ over a pool
$S = \{z_i\}_{i=1}^{n}$, where $w_i$ multiplies the loss of example $i$ during
training and $w_i = 0$ excludes it. Let $\mathcal{A}$ map a weighting to trained
parameters, let $\phi$ measure a trained model, and let
$f = \phi \circ \mathcal{A}$. Evaluating $f$ requires retraining, so attribution
replaces it with an explicit $\hat{f}$, the first-order expansion about the full
dataset $\mathbf{1}_{n}$:
\begin{equation}
  \label{eq:app-influence}
  \hat{f}(\mathbf{w}) := f(\mathbf{1}_{n})
    + \left( \frac{\partial f(\mathbf{w})}{\partial \mathbf{w}}
        \bigg|_{\mathbf{w} = \mathbf{1}_{n}} \right)^{\!\top}
      (\mathbf{w} - \mathbf{1}_{n}).
\end{equation}
The leading term is the \emph{influence function}
\citep{koh2017understanding,ilyas2025magic}. In deep learning, it has no closed
form, because the derivative runs through training itself, so scalable methods
approximate it (\refsec{related}). MAGIC computes it exactly.

\paragraph{Single-model attribution.}
Training is stochastic, so conventional attribution predicts what a freshly
initialized model would do on average without an example, and some of the
variation in $f$ does not come from the data. Fixing initialization and data
order removes that variation: $\phi(\mathcal{A}(\mathbf{w}))$ becomes
deterministic, and model output varies with the data weights alone
\citep{ilyas2025magic}. Our retraining checks resume from a stored optimizer
state in the same data order, so they measure the quantity the scores predict.
For the same reason, a score describes only the run it was computed on, and
block rankings need not carry over to another seed (\refsec{families}).

\paragraph{Iterative and smooth training.}
\textsc{Replay} computes the exact metagradient when training is iterative and
smooth \citep{ilyas2025magic}. Iterative means
\begin{equation}
  \label{eq:app-iterative}
  \mathcal{A}(\mathbf{w}) := s_{T}
  \quad \text{for} \quad
  s_{t+1} := h_t\!\left( s_t, \mathbf{g}_t(s_t, \mathbf{w}) \right)
  \quad \text{and} \quad
  \mathbf{g}_t(s_t, \mathbf{w}) := \sum_{i \in B_t} w_i \cdot \nabla_{s_t} \ell(z_i; s_t),
\end{equation}
where $s_t$ is the optimizer state at step $t$, $h_t$ the update rule,
$B_t \subseteq [n]$ the minibatch, $\ell(z_i; s_t)$ the loss of example $z_i$ at
that state, and $T$ the number of steps. In a window, $s_0$ is the exported
checkpoint and $T$ is the window length. Smooth means that for any measurement
function $\phi$, a small change in $\mathbf{w}$ makes only a small change in
$\nicefrac{\partial f(\mathbf{w})}{\partial \mathbf{w}}$;
\citet{engstrom2025optimizing} call this metasmoothness.

For AdamW \citep{loshchilov2019decoupled}, $s_t = (\theta_t, m_t, v_t)$ holds the parameters and the first and
second moment estimates. The update is not smooth as usually written, because the
derivative of $\sqrt{v_t}$ grows without bound as $v_t$ approaches zero. The
smooth version replaces that root with $\sqrt{v_t + \epsilon_{\mathrm{root}}}$
for a constant $\epsilon_{\mathrm{root}} > 0$; the update we differentiate is the
one implemented in \texttt{bergson} \citep{quirke2026bergson}. This constant is
the damping term of \reftab{envelope} and \reftab{damping}. At $86$M,
$\epsilon_{\mathrm{root}} = 0$ gives non-finite scores and $10^{-6}$ gives LDS
$0.96$, while at $10^{-4}$ or $10^{-2}$ the model no longer learns arithmetic.
Smoothness comes from the training configuration, which is the formal basis for
the finding in \refsec{windows} that attribution validity depends on the
training recipe.

\paragraph{Exact computation.}
The gradient of $f$ with respect to $\mathbf{w}$ is a metagradient, so
\textsc{Replay} \citep{engstrom2025optimizing} computes it by walking the
realized training path backward. It initializes
$\Delta_T = \nabla_{s_T} \phi(s_T)$; then for $t = T-1, \ldots, 0$ it reloads
$s_t$ and $B_t$ and computes
\begin{equation*}
  \beta_t = \nabla_{\mathbf{w}} \!\left( h_t(s_t, \mathbf{g}_t(s_t, \mathbf{w}))^{\top} \Delta_{t+1} \right),
  \qquad
  \Delta_t = \nabla_{s_t} \!\left( h_t(s_t, \mathbf{g}_t(s_t, \mathbf{w}))^{\top} \Delta_{t+1} \right),
\end{equation*}
where $\beta_t$ is the part of the derivative that arises at step $t$, and
$\Delta_t = \nicefrac{\partial f(\mathbf{w})}{\partial s_t}$ is the sensitivity
of the target to the optimizer state at that step. The exact influence function
is $\beta = \sum_{t=0}^{T-1} \beta_t$. \textsc{Replay} keeps a small set of
states and recomputes the others, which costs $T + T \log T$ training steps and
$\log T$ memory \citep{ilyas2025magic}. Because a score sums over every step of
the differentiated window, it is defined only for a named window. And
$\Delta_t$ is a single running quantity carried through all $T$ steps without
normalization, which is why it overflows over a full run (\refsec{fullrun}).

\paragraph{Validation: the linear datamodeling score.}
Because $\hat{f}$ predicts a retraining outcome, retraining can check it. Sample
$M$ binary weight vectors $\mathbf{w}^{(1)}, \ldots, \mathbf{w}^{(M)} \in \{0,1\}^{n}$,
each obtained by dropping $pn$ blocks at random for a drop fraction
$p \in [0,1]$. For each vector, obtain the true
$f(\mathbf{w}^{(j)})$ by retraining and the predicted $\hat{f}(\mathbf{w}^{(j)})$
from \eqref{eq:app-influence}, then report the Spearman correlation between the
two:
\begin{equation}
  \label{eq:app-lds}
  \mathrm{LDS} = \rho\!\left(
    \left[ \hat{f}(\mathbf{w}^{(j)}) \right]_{j=1}^{M},\;
    \left[ f(\mathbf{w}^{(j)}) \right]_{j=1}^{M}
  \right).
\end{equation}
Subsets are indexed by superscript, as in $\mathbf{w}^{(j)}$, and the weight on
block $i$ by subscript, as in $w_i$, so $M$ counts retrains and $n$ counts
blocks. The four main $152$M windows use $p = 0.01$ and $M = 48$ subsets,
computed per held-out query and aggregated across eight queries. Full-run
diagnostics use $12$ subsets; the statistical $p$-values we report for them are
separate from the removal fraction $p$. \reftab{window-index} lists the other
window configurations. \citet{ilyas2025magic} report that LDS degrades as $p$
grows, because the linear estimate holds only near $\mathbf{1}_{n}$. Our failures
come instead from the horizon $T$ at fixed $p$.

\paragraph{How this study applies the framework.}
The pool $S$ is the set of $512$-token blocks the differentiated window
consumes, so $\mathbf{w}$ indexes block occurrences, and $n$ counts blocks in the
window rather than in the corpus. The measurement $\phi$ is mean completion-only
negative log-likelihood on the held-out programs, evaluated at the final state
of the window. For a window, $\mathcal{A}$ continues from an exported
checkpoint $s_0$. The four main $152$M windows use $T = 100$ AdamW steps in fp32
at $512$ rows per step, on blocks the parent run had not yet reached.

The sign convention of \refsec{method} follows from \eqref{eq:app-influence}.
The on-disk score of block $i$ is $\beta_i$, the $i$-th component of the $\beta$
that \textsc{Replay} returns, that is
$\nicefrac{\partial f(\mathbf{w})}{\partial w_i}$ at $\mathbf{w} = \mathbf{1}_{n}$.
Removing a subset $R$ sets $w_i = 0$ for $i \in R$, so
$(\mathbf{w} - \mathbf{1}_{n})_i = -1$ on $R$ and zero elsewhere, and
\begin{equation}
  \label{eq:app-sign}
  \hat{f}(\mathbf{w}) - f(\mathbf{1}_{n}) = -\sum_{i \in R} \beta_i .
\end{equation}
A positive score therefore predicts that removing the block \emph{lowers}
held-out loss: positive means harmful, and the most helpful blocks carry the most
negative scores. Equivalently, $\sum_{i \in R} \beta_i$ predicts baseline
held-out loss minus the loss after removing $R$, which our validation harness
correlates with measured retraining at a median of $0.86$--$0.99$ per window
(\refapp{attribution}).

\end{document}